\documentclass[cameraready]{Interspeech} 

\title{When Multiple Scripts Matter: \\Evaluating ASR in Clinical Settings}

\author[affiliation={1, 2}, equalcontribution]{Jean}{Seo}
\author[affiliation={1}, equalcontribution]{Minkyu}{Kim}
\author[affiliation={1}]{Jeonguk}{Lee}
\author[affiliation={1}]{Jisoo}{Jung}
\author[affiliation={1}]{Wooseok}{Han}
\author[affiliation={1,3}, correspondingauthor]{Eunho}{Yang}

\address{
    $^1$ AITRICS, $^2$ University of Copenhagen, $^3$ KAIST
}

\email{jean.seo@di.ku.dk\\ minkyu.kim@aitrics.com}


\usepackage{comment}
\usepackage{CJKutf8}
\usepackage{tabularx}
\usepackage{array}
\usepackage{etoolbox}
\usepackage{booktabs}
\usepackage{multirow}
\usepackage{array}
\usepackage{fix-cm}

\usepackage{booktabs}
\usepackage[table]{xcolor} 
\usepackage{siunitx} 

\usepackage{soul}
\usepackage{xcolor}

\usepackage{algorithm}
\usepackage{algorithmic}

\definecolor{edit}{RGB}{0, 0, 0} 

\begin{document}

\maketitle

\begin{abstract}

Automatic speech recognition (ASR) in non-English clinical settings is challenged by multiscript variability, where the same term may appear in multiple valid orthographic forms. Conventional string-matching evaluation metrics often underestimate ASR performance by treating orthographic variants as errors. To address this issue, we introduce \textbf{MultiClin}, a clinical ASR benchmark designed to evaluate robustness to multiscript variability. Experiments across diverse ASR models show that multiscript-aware evaluation provides a fairer assessment of recognition quality than conventional single-reference evaluation. We further investigate the impact of script consistency during training and find that inconsistent script mappings increase orthographic uncertainty and hinder model convergence, with a balanced 50\% mapping ratio producing the highest entropy. In contrast, script unification consistently yields the best ASR performance. Our dataset and code are publicly available at: \url{https://github.com/aitrics-ronaldo/Interspeech_MultiClin}.

\end{abstract}

\section{Introduction}

Automatic speech recognition (ASR) is increasingly adopted in clinical settings to improve workflow efficiency \cite{xu2025enhancing, alboksmaty2025impact, tran2023automatic}. However, domain-specific terminology and noisy environments continue to challenge clinical ASR. These difficulties are further amplified in non-English settings, where English medical terminology frequently coexists with phonetic renderings in local scripts \cite{agro2025codeswitchingendtoendautomaticspeech}. A central obstacle to reliable benchmarking in such environments is \emph{multiscript variability}, where a single spoken term may correspond to multiple valid orthographic forms (e.g., English spelling or a phonetic rendering in the local script). Unlike conventional code-switching, which involves acoustic alternation between languages, multiscript variability arises from orthographic variation despite an identical acoustic realization.

Conventional ASR evaluation assumes a single reference transcription per utterance. However, this assumption often breaks down in non-English clinical settings, where English-origin medical terms lack standardized localization guidelines and may be transcribed in multiple valid forms. This many-to-one mapping between orthography and speech invalidates strict string-based metrics such as word error rate (WER), systematically penalizing outputs that are phonetically and semantically correct but orthographically different from the reference \cite{Mustafa2022CodeSwitchingIA, Srivastava2018HomophoneIA, Chowdhury2020EffectsOD}. Moreover, normalization-based solutions remain impractical due to inconsistent clinical documentation practices and the scarcity of standardized domain-specific corpora. While multilingual ASR research has extensively studied code-switching \cite{Nakayama2019ZeroShotCA}, prior work has largely focused on modeling and data augmentation \cite{kumar2021dual, Li2019TowardsCA, Yilmaz2017LanguageDF} rather than evaluation. Existing benchmarks typically rely on a single ground-truth reference \cite{Hamed2022BenchmarkingEM, Paik2025HiKEHE}, while transliteration-based approaches \cite{Emond2018TransliterationBA} and metrics such as transliterated WER (T-WER) \cite{Chowdhury2021TowardsOM, Ali2015MultiReferenceEF} have primarily been evaluated on general-domain code-switching and dialectal variation, leaving clinical multiscript settings largely unexplored.

\begin{table}[t]
\centering
\begingroup
\fontsize{6.5}{7.5}\selectfont
\setlength{\tabcolsep}{3pt}
\renewcommand{\arraystretch}{1.3}
\caption{Example of the original, tagged, and translated dialogue from the \textbf{MultiClin} dataset.} \label{tab:example_summarized}
\begin{tabular}{|>{\raggedright\arraybackslash}p{\dimexpr\linewidth-2\tabcolsep-2\arrayrulewidth\relax}|}
\hline
\textbf{Doctor-Patient Dialogue} \\
\hline
\textbf{Original} \\
{[}patient{]} How long do I have to wear that brace? \\
{[}doctor{]} You're gonna be wearing the brace for about 6 weeks. \\
{[}patient{]} 6 weeks? \\
... \\ \hline
\textbf{Tagged} \\
{[}patient{]} How long do I have to wear that \textless{}MEDICAL\textgreater{}brace\textless{}/MEDICAL\textgreater{}? \\
{[}doctor{]} You're gonna be wearing the \textless{}MEDICAL\textgreater{}brace\textless{}/MEDICAL\textgreater{} for about \textless{}NUMBER\textgreater{}6\textless{}/NUMBER\textgreater{} weeks. \\
{[}patient{]} \textless{}NUMBER\textgreater{}6\textless{}/NUMBER\textgreater{} weeks? \\
... \\ \hline
\textbf{Translated} \\
{[}patient{]} \begin{CJK}{UTF8}{mj}그\end{CJK} \textless{}MEDICAL\textgreater{}brace,\begin{CJK}{UTF8}{mj}브레이스\end{CJK}\textless{}/MEDICAL\textgreater{}\begin{CJK}{UTF8}{mj}는 얼마나 오래 착용해야 하나요?\end{CJK} \\
{[}doctor{]} \textless{}MEDICAL\textgreater{}brace,\begin{CJK}{UTF8}{mj}브레이스\end{CJK}\textless{}/MEDICAL\textgreater{}\begin{CJK}{UTF8}{mj}는 약\end{CJK} \textless{}NUMBER\textgreater{}6,\begin{CJK}{UTF8}{mj}육\end{CJK}\textless{}/NUMBER\textgreater{}\begin{CJK}{UTF8}{mj}주 정도 착용하시게 될 거예요\end{CJK}. \\
{[}patient{]} \textless{}NUMBER\textgreater{}6,\begin{CJK}{UTF8}{mj}육\end{CJK}\textless{}/NUMBER\textgreater{}\begin{CJK}{UTF8}{mj}주요\end{CJK}? \\
... \\ \hline
\end{tabular}
\endgroup
\end{table}

To address this gap, we introduce \textbf{MultiClin}, a clinical ASR benchmark that provides multiple valid transcription variants for multiscript terminology. Through a Korean clinical case study, we demonstrate that dynamic multi-reference evaluation yields a fairer assessment of ASR performance under orthographic variability.

\begin{table*}[!t]
\centering
\small
\renewcommand{\arraystretch}{1.0}

\caption{Statistics of the \textbf{MultiClin} dataset.} 
\label{tab:dataset_summary}

\begin{tabular*}{\textwidth}{l r @{\extracolsep{\fill}} l r}
\toprule
\multicolumn{2}{c}{\textbf{A. Filtering Stages (Initial $\to$ Final)}} & 
\multicolumn{2}{c}{\textbf{B. Avg. tagged instances per dialogue}} \\
\midrule
ACI Bench      & $126 \to 116$ & \textsc{Medical} Tags & 44 \\
Primock57       & $186 \to 9$   & \textsc{Number} Tags  & 6  \\
MTS-Dialog      & $1,175 \to 191$ & \textsc{Unit} Tags    & 1  \\
\textbf{Total Dialogues} & \textbf{$1,487 \to 316$} & & \\
\bottomrule
\end{tabular*}

\vspace{1.5em}

\begin{tabular*}{\textwidth}{l r @{\extracolsep{\fill}} l r}
\toprule
\multicolumn{2}{c}{\textbf{C. Avg. Utterances Per Dialogue}} & 
\multicolumn{2}{c}{\textbf{D. \# Speaker composition}} \\
\midrule
Doctor          & 18.0 & Doctor, Patient & 268 \\
Doctor 2         & 0.3  & Doctor, Patient, Guest\_Family & 22 \\
Patient         & 16.0 & Doctor, Guest\_Family & 11 \\
Guest\_Family   & 0.9  & Doctor, Doctor 2, Patient & 9 \\
Guest\_Family 2  & 0.03 & Other & 6 \\
\bottomrule
\end{tabular*}
\end{table*}




\section{MultiClin dataset}

We construct the \textbf{MultiClin} dataset to reflect real-world clinical ASR challenges. Table \ref{tab:example_summarized} illustrates an example data corresponding to each phase of the annotation process.






\subsection{Dataset construction}
\subsubsection{Collection}

We collect publicly available doctor–patient dialogues from ACIBench \cite{yim2023acibenchnovelambientclinical}, Primock57 \cite{papadopoulos-korfiatis-etal-2022-primock57}, and MTS-Dialog \cite{mts-dialog}. To ensure natural clinical conversations, we exclude dialogues involving virtual assistants and retain only interactions between doctors and patients, resulting in an initial corpus of 1,487 dialogues.

\subsubsection{Annotation}
\label{section:annotation}


The dataset undergoes three processing stages: tagging, translation, and human annotation. We use gpt-5.2\footnote{\url{https://openai.com/}} to identify script-switching instances and assign them to three categories: \textsc{Medical}, \textsc{Unit}, and \textsc{Number}. \textsc{Medical} tags denote English-origin medical terms appearing either in the Roman alphabet or as phonetic loanwords. \textsc{Unit} tags represent measurement units expressed in native scripts or standardized symbols (e.g., \%, cm), while \textsc{Number} tags capture numerical expressions written in native scripts or Arabic numerals. We then translate the dialogues into Korean using the same model. Tagged spans preserve their original form and are augmented with Korean-script (Hangeul) renderings, separated by commas without spaces. For example, ``You need an \textless medical\textgreater injection\textless/medical\textgreater.'' becomes ``\textless medical\textgreater injection,\begin{CJK}{UTF8}{mj}인젝션\end{CJK}\textless/medical\textgreater \begin{CJK}{UTF8}{mj}이 필요합니다\end{CJK}.'' In other words, tagged entities undergo transliteration, preserving lexical identity while changing only the script, whereas the remaining text undergoes full translation into Korean. Finally, two annotators with nursing backgrounds review all dialogues for orthographic correctness, translation fidelity, and naturalness. Any disagreements or errors are resolved through consensus, resulting in the final curated dataset.


\subsubsection{Speech generation}



To comply with the Health Insurance Portability and Accountability Act (HIPAA) restrictions on releasing real-world clinical audio, we synthesize dialogues using gpt-4o-mini-tts. We map speaker roles to distinct speaking styles (e.g., professional tones for doctors and lethargic tones for patients) and apply accent-aware prompting to align multiscript spans with native intonation patterns. To reduce the acoustic mismatch between synthetic and real clinical speech, we incorporate human-like conversational dynamics, including overlaps and response latencies, and simulate clinical environments using a DSP chain \footnote{\url{https://github.com/spotify/pedalboard}} (e.g., reverb and HVAC noise). All audio is resampled to $16,\text{kHz}$.

\begin{table}[h t]
\centering
\small
\caption{Clinical specialty distribution in \textbf{MultiClin}. \textit{Other}$^{\ast}$ encompasses 8 minor fields (e.g., Pain Management, Dentistry, Plastic Surgery).}
\label{tab:specialty_distribution}
\begin{tabular*}{\columnwidth}{@{\extracolsep{\fill}} p{0.52\columnwidth} rr @{}}
\toprule
\textbf{Primary Specialty} & \textbf{\# Dialogues} & \textbf{Ratio} \\
\midrule
Orthopedics & 95 & 0.301 \\
Family Medicine / General Practice & 34 & 0.108 \\
Neurology & 30 & 0.095 \\
Urology & 19 & 0.060 \\
Cardiology & 18 & 0.057 \\
Pulmonology & 16 & 0.051 \\
Gastroenterology & 12 & 0.038 \\
Hematology / Oncology & 9 & 0.028 \\
ENT (Otolaryngology) & 8 & 0.025 \\
Rheumatology & 7 & 0.022 \\
Dermatology & 7 & 0.022 \\
Psychiatry & 7 & 0.022 \\
Pediatrics & 6 & 0.019 \\
Neurosurgery & 5 & 0.016 \\
Emergency Medicine & 5 & 0.016 \\
General Surgery & 5 & 0.016 \\
Ophthalmology & 4 & 0.013 \\
Obstetrics \& Gynecology & 4 & 0.013 \\
Allergy / Immunology & 4 & 0.013 \\
Endocrinology & 3 & 0.009 \\
Other$^{\ast}$ & 14 & 0.044 \\
\bottomrule
\end{tabular*}
\end{table}

\subsection{Statistics}

\noindent\textbf{Dataset filtering.} From the initial 1,487 dialogues, we retain 1,417 instances containing at least one \textsc{Medical}, \textsc{Number}, or \textsc{Unit} tag. We then manually remove unnatural or hallucinated conversations, resulting in 316 final dialogues (Table~\ref{tab:dataset_summary}A).

\smallskip
\noindent\textbf{Tag and dialogue statistics.} \textsc{Medical} terminology dominates script-switching instances (Table~\ref{tab:dataset_summary}B). Each dialogue contains 34 turns and 68 sentences on average, with per-speaker utterance statistics reported in Table~\ref{tab:dataset_summary}C.

\smallskip
\noindent\textbf{Speaker composition.} Most dialogues involve a single doctor and a single patient (Table~\ref{tab:dataset_summary}D). In cases where a patient is absent, guardians (\textsc{Guest Family}) speak on their behalf.

\smallskip
\noindent\textbf{Clinical specialty distribution.} All dialogues across the three sources are accompanied by structured clinical notes (e.g., SOAP notes). Using gpt-5.2, we infer the primary clinical specialty of each dialogue from this metadata (Table~\ref{tab:specialty_distribution}).

\begin{table*}[!t]
\centering
\footnotesize
\renewcommand{\arraystretch}{1.1} 
\setlength{\tabcolsep}{0pt}   
\caption{Performance of baseline models.} \label{tab:pre_trained_results}
\begin{tabular*}{\textwidth}{@{\extracolsep{\fill}} lll cc cc cc cc cc cc @{}}
\toprule
\multirow{2}{*}{\textbf{Medical}} & \multirow{2}{*}{\textbf{Number}} & \multirow{2}{*}{\textbf{Unit}} & \multicolumn{2}{c}{\textbf{Whisper v3}} & \multicolumn{2}{c}{\textbf{v3 Turbo}} & \multicolumn{2}{c}{\textbf{Qwen3 0.6B}} & \multicolumn{2}{c}{\textbf{Qwen3 1.7B}} & \multicolumn{2}{c}{\textbf{Gemini 2.5 Flash}} & \multicolumn{2}{c}{\textbf{Gemini 2.5 Pro}} \\
\cmidrule(lr){4-5} \cmidrule(lr){6-7} \cmidrule(lr){8-9} \cmidrule(lr){10-11} \cmidrule(lr){12-13} \cmidrule(lr){14-15}
& & & \textbf{CER} & \textbf{WER} & \textbf{CER} & \textbf{WER} & \textbf{CER} & \textbf{WER} & \textbf{CER} & \textbf{WER} & \textbf{CER} & \textbf{WER} & \textbf{CER} & \textbf{WER} \\ 
\midrule
\multirow{4}{*}{original} & \multirow{2}{*}{original} & original & 29.68 & 36.57 & 26.65 & 32.44 & 59.09 & 83.09 & 29.73 & 41.83 & 24.87 & 30.83 & 24.23 & 28.28 \\
& & both & 29.51 & 36.56 & 26.46 & 32.42 & 58.85 & 83.06 & 29.38 & 41.79 & 24.66 & 30.79 & 24.02 & 28.24 \\ \cmidrule(lr){2-15}
& \multirow{2}{*}{both} & original & 29.83 & 36.76 & 26.80 & 32.62 & 59.05 & 83.08 & 29.69 & 41.81 & 25.01 & 31.00 & 24.37 & 28.47 \\
& & both & 29.65 & 36.75 & 26.60 & 32.61 & 58.80 & 83.04 & 29.34 & 41.76 & 24.80 & 30.98 & 24.16 & 28.44 \\ \midrule
\multirow{4}{*}{both} & \multirow{2}{*}{original} & original & 13.37 & 27.78 & 8.71 & 22.89 & 48.77 & 80.14 & 15.12 & 37.11 & 6.01 & 19.16 & 5.06 & 15.67 \\
& & both & 13.12 & 27.76 & 8.44 & 22.87 & 48.46 & 80.11 & 14.68 & 37.06 & 5.72 & 19.12 & 4.76 & 15.64 \\ \cmidrule(lr){2-15}
& \multirow{2}{*}{both} & original & 13.48 & 27.92 & 8.82 & 23.02 & 48.69 & 80.10 & 15.04 & 37.06 & 6.12 & 19.29 & 5.15 & 15.81 \\
& & both & 13.24 & 27.91 & 8.55 & 23.00 & 48.38 & 80.07 & 14.60 & 37.01 & 5.83 & 19.27 & 4.86 & 15.78 \\ 
\bottomrule
\end{tabular*}
\end{table*}

\section{Experiments}
\label{section:experiments}

We evaluate ASR performance on the \textbf{MultiClin} benchmark to quantify the impact of multiscript variability. We analyze zero-shot inference across diverse architectures and assess the effects of domain-specific fine-tuning under different labeling strategies.

\subsection{Experimental setup}

\subsubsection{Baseline Models}

We consider three model families as baselines: (1) \textbf{Whisper} \cite{whisper} (\textit{large-v3}, \textit{v3-turbo}), implemented via faster-whisper\footnote{\url{https://github.com/SYSTRAN/faster-whisper}}; (2) \textbf{Qwen3 ASR} \cite{qwen3-asr} (\textit{0.6B}, \textit{1.7B}); and (3) \textbf{Gemini} \cite{gemini} (\textit{2.5 Flash}, \textit{2.5 Pro}), representing frontier multimodal state-of-the-art models.




\subsubsection{Inference Configuration}
We detail the zero-shot inference configurations for our multimodal baselines to ensure reproducibility.

\smallskip
\noindent\textbf{Gemini prompting strategy.} We query the Gemini models using a structured zero-shot prompt. We instruct the model to act as a professional medical stenographer and produce verbatim transcriptions, explicitly prohibiting speaker diarization, speaker prefixes, and summarization. To ensure deterministic and parseable outputs, we set the sampling temperature to 0.0 and enforce a JSON output format, from which we extract the transcript as an array of sentences.

\smallskip
\noindent\textbf{Qwen inference setting.} For Qwen3 ASR models, we accommodate long clinical dialogues by setting the maximum generation length to 65,536 tokens. To improve memory efficiency and avoid out-of-memory (OOM) errors during long-form audio processing, we limit the maximum inference batch size to 32.


\subsubsection{Fine-tuning Configuration}
For the fine-tuning experiments, we train Whisper models using LoRA \cite{lora}. We split the \textbf{MultiClin} dataset into a 9:1 ratio to construct an independent test set. Importantly, we apply a 100\% transliteration ratio, in which all tagged \textsc{Medical}, \textsc{Number}, and \textsc{Unit} entities are consistently unified into the local script to maximize labeling consistency. This setup reduces orthographic ambiguity during the learning phase. Finally, models are trained for 4 epochs with a batch size of 4.

\subsubsection{Evaluation Protocol}
To enable more accurate evaluation of ASR performance under multiscript settings, we introduce a localized evaluation metric (Algorithm~\ref{alg:dynamic_ref}) that treats both the original English medical term and its phonetic rendering in the local script as valid references. Specifically, for each script-switching entity in the reference transcript, we dynamically extract a 50-character window from the ASR prediction $\mathbf{\hat{y}}$ using a tracking cursor. To mitigate temporal misalignment, we apply Longest Common Substring (LCS) matching between the target entity and the corresponding predicted window. We then compute local CER and WER within these aligned boundaries, reducing the influence of surrounding transcription errors and enabling a more robust comparison of entity-level correctness across orthographic variants.

\begin{algorithm}[!t]
\caption{Dynamic Multiscript Reference Resolution}
\label{alg:dynamic_ref}
\textbf{Input:} Tagged reference $\mathbf{y}_{tag}$, ASR hypothesis $\mathbf{\hat{y}}$, Window size $W=50$, Mode mapping $\mathcal{M} \in \{\text{original}, \text{both}\}$ \\
\noindent\textbf{Output:} \raggedright Dynamically resolved reference $\mathbf{y}_{final}$\par
\begin{algorithmic}[1]
\STATE $cursor \leftarrow 0$
\STATE $\mathbf{y}_{final} \leftarrow \mathbf{y}_{tag}$
\FOR{each entity tuple $(t, e_{orig}, e_{tgt})$ in $\mathbf{y}_{tag}$}
    \STATE $m \leftarrow \mathcal{M}[t]$ \COMMENT{Fetch evaluation mode for tag type $t$}
    
    \IF{$m = \text{original}$}
        \STATE Replace tag with $e_{orig}$ in $\mathbf{y}_{final}$
    \ELSIF{$m = \text{both}$}
        \IF{$cursor \geq |\mathbf{\hat{y}}|$}
            \STATE Replace tag with $e_{orig}$ in $\mathbf{y}_{final}$
        \ELSE
            \STATE $\mathbf{\hat{y}}_{win} \leftarrow \mathbf{\hat{y}}[cursor : \min(cursor + W, |\mathbf{\hat{y}}|)]$
            \STATE $cer_{orig}, offset_{orig} \leftarrow \text{LocalCER}(e_{orig}, \mathbf{\hat{y}}_{win})$
            \STATE $cer_{tgt}, offset_{tgt} \leftarrow \text{LocalCER}(e_{tgt}, \mathbf{\hat{y}}_{win})$
            
            \STATE \COMMENT{Priority selection based on minimal local error}
            \IF{$cer_{tgt} < cer_{orig}$}
                \STATE Replace tag with $e_{tgt}$ in $\mathbf{y}_{final}$
                \STATE $cursor \leftarrow cursor + offset_{tgt}$
            \ELSE
                \STATE Replace tag with $e_{orig}$ in $\mathbf{y}_{final}$
                \STATE $cursor \leftarrow cursor + offset_{orig}$
            \ENDIF
        \ENDIF
    \ENDIF
\ENDFOR
\RETURN $\mathbf{y}_{final}$

\vspace{0.1cm}
\hrule
\vspace{0.1cm}

\STATE \textbf{Function} $\text{LocalCER}(e, \mathbf{w})$
\STATE \quad $\text{LCS} \leftarrow \text{FindLongestMatch}(e, \mathbf{w})$
\STATE \quad $\mathbf{w}_{sub} \leftarrow \mathbf{w}[\text{LCS}_{start} : \text{LCS}_{end}]$
\STATE \quad $cer \leftarrow \text{ComputeCER}(e, \mathbf{w}_{sub})$
\STATE \quad \textbf{return} $cer, \text{LCS}_{end}$
\end{algorithmic}
\end{algorithm}



\subsection{Inference results}

Table \ref{tab:pre_trained_results} presents zero-shot inference performance across different script evaluation settings. A consistent trend emerges: moving from strict single-label matching (\textit{original}) to multiscript-aware evaluation (\textit{both}) yields substantial reductions in error rates across all models. For instance, Gemini 2.5 Pro's WER decreases from 28.28\% to 15.78\% when medical terms are evaluated with multiscript flexibility. These results empirically support the claim that conventional string-based metrics systematically underestimate ASR performance by failing to account for valid orthographic variation in clinical settings. Our proposed benchmark exposes this limitation by revealing the true capabilities of ASR models. While \textsc{Medical} tags contribute most to the observed performance gap, model scale also plays a significant role; Qwen3 ASR 1.7B achieves a 37.01\% WER under the full multiscript-aware setting (\textit{both}). Among open-source systems, Whisper v3 Turbo demonstrates the strongest robustness, achieving a 23.00\% WER. Overall, Gemini 2.5 Pro attains the best CER of 4.86\%. By properly accounting for medical-domain orthographic variation, \textbf{MultiClin} provides a more fair and informative evaluation framework for multiscript clinical ASR.

\begin{table}[ht]
\centering
\footnotesize
\renewcommand{\arraystretch}{1.3} 
\setlength{\tabcolsep}{0pt}
\caption{Detailed CER (\%) comparison between pre-trained and fine-tuned Whisper models. Parentheses indicate the absolute reduction in CER after fine-tuning on the \textbf{MultiClin} dataset.} \label{tab:main_fine_tuned}
\begin{tabular*}{\columnwidth}{@{\extracolsep{\fill}} llllcc @{}}
\hline
\multirow{2}{*}{\scriptsize\textbf{Models}} & \multirow{2}{*}{\scriptsize\textbf{Medical}} & \multirow{2}{*}{\scriptsize\textbf{Number}} & \multirow{2}{*}{\scriptsize\textbf{Unit}} & \scriptsize\textbf{Pretrained} & \scriptsize\textbf{Finetuned} \\ \cline{5-6} 
 &  &  &  & \scriptsize\textbf{CER (\%)} & \scriptsize\textbf{CER (\%)} \\ \hline
\multirow{8}{*}{\shortstack{Large v3}} & \multirow{4}{*}{original} & \multirow{2}{*}{original} & original & 30.18 & 27.84 {\tiny($\downarrow$2.34)} \\
 &  &  & both & 30.06 & 27.53 {\tiny($\downarrow$2.53)} \\ \cline{3-6} 
 &  & \multirow{2}{*}{both} & original & 30.35 & 27.58 {\tiny($\downarrow$2.77)} \\
 &  &  & both & 30.24 & 27.28 {\tiny($\downarrow$2.96)} \\ \cline{2-6} 
 & \multirow{4}{*}{both} & \multirow{2}{*}{original} & original & 14.01 & 8.39 {\tiny($\downarrow$5.62)} \\
 &  &  & both & 13.85 & 8.00 {\tiny($\downarrow$5.85)} \\ \cline{3-6} 
 &  & \multirow{2}{*}{both} & original & 14.16 & 8.08 {\tiny($\downarrow$6.08)} \\
 &  &  & both & 13.99 & 7.66 {\tiny($\downarrow$6.33)} \\ \hline
\multirow{8}{*}{\shortstack{Large v3 \\ Turbo}} & \multirow{4}{*}{original} & \multirow{2}{*}{original} & original & 28.73 & 26.77 {\tiny($\downarrow$1.96)} \\
 &  &  & both & 28.59 & 26.47 {\tiny($\downarrow$2.12)} \\ \cline{3-6} 
 &  & \multirow{2}{*}{both} & original & 28.91 & 26.55 {\tiny($\downarrow$2.36)} \\
 &  &  & both & 28.76 & 26.23 {\tiny($\downarrow$2.53)} \\ \cline{2-6} 
 & \multirow{4}{*}{both} & \multirow{2}{*}{original} & original & 10.08 & 6.85 {\tiny($\downarrow$3.23)} \\
 &  &  & both & 9.88 & 6.47 {\tiny($\downarrow$3.41)} \\ \cline{3-6} 
 &  & \multirow{2}{*}{both} & original & 10.19 & 6.58 {\tiny($\downarrow$3.61)} \\
 &  &  & both & 9.99 & \textbf{6.16 {\tiny($\downarrow$3.83)}} \\ \hline
\end{tabular*}
\end{table}
\begin{table}[!ht]
\centering
\scriptsize
\renewcommand{\arraystretch}{1.1}
\caption{Impact of transliteration ratio in the training dataset. Results show the performances of fine-tuned Whisper large v3 models on the held-out test set.} \label{tab:labeling_consistency}
\begin{tabular}{lcc}
\toprule
\textbf{Ratio (\%)} & \textbf{CER (\%)} & \textbf{WER (\%)} \\ 
\midrule
0   & 69.17 & 54.35 \\ 
25  & 27.42 & 30.51 \\ 
\textbf{50}  & \textcolor{red}{57.47} & \textcolor{red}{48.50} \\ 
75  & 13.53 & 22.55 \\ 
\textbf{100} & \textbf{7.66} & \textbf{17.48} \\ 
\bottomrule
\end{tabular}
\end{table}



\subsection{Fine-tuning results}

Table \ref{tab:main_fine_tuned} summarizes the performance gains from fine-tuning on the independent test set. Training with a 100\% transliteration ratio yields substantial improvements across all evaluation settings. Notably, Whisper-Large v3 Turbo achieves a best-in-class CER of 6.16\%, corresponding to an absolute reduction of 3.83\%p over its pre-trained baseline. Even larger gains are observed for the standard Whisper-Large v3 model, with CER decreasing by up to 6.33\%p under multiscript-aware evaluation criteria. These consistent improvements across both architectures empirically demonstrate that full script unification is an effective strategy for mitigating orthographic ambiguity in clinical ASR.

\subsection{Impact of labeling consistency}

We further investigate the impact of labeling consistency in the training data. As shown in Table \ref{tab:labeling_consistency}, the 0\% transliteration ratio—where all tagged entities are represented exclusively in the Roman alphabet or Arabic numerals while the rest of the utterance is written in Korean (Hangeul)—produces the highest error rates on the held-out test set (69.17\% CER and 54.35\% WER). As the transliteration ratio increases, meaning that a larger proportion of entities are represented in Hangeul, performance exhibits a non-monotonic pattern, with a secondary error peak at the 50\% ratio (57.47\% CER and 48.50\% WER). This performance degradation confirms that inconsistent script mapping introduces orthographic ambiguity, maximizing the conditional entropy $H(Y|X)$ for a given acoustic feature $X$:
$$H(Y|X) = -\sum_{y \in \mathcal{Y}} P(y|X) \log P(y|X)$$
At the 50\% ratio, the model faces maximum epistemic uncertainty between competing scripts, which disrupts internal alignment and prevents the decoder from forming stable decision boundaries. Ultimately, the 100\% ratio resolves the script alternation complexity, yielding the most robust performance (7.66\% CER, 17.48\% WER). This validates that full script unification is essential for providing a deterministic learning signal.

\section{Conclusion}

This work introduces the \textbf{MultiClin} dataset for fairer evaluation in non-English clinical ASR. Our experiments show that multiscript-aware criteria provide a fairer assessment than traditional single-label metrics, which often underestimate true model performance. We further demonstrate that labeling consistency in the training data is essential for better performance. Future work should examine how these ASR improvements influence downstream clinical tasks, such as entity extraction and SOAP note generation.



\section{Generative AI Use Disclosure}
This work employs Generative AI tools including Google Gemini and OpenAI ChatGPT. Gemini is utilized for linguistic refinement, including grammatical correction and improving the clarity of the initial manuscript. Furthermore, both Gemini and ChatGPT were integrated into our data construction process to generate synthetic clinical dialogues for the \textbf{MultiClin} dataset, addressing the inherent data scarcity and privacy constraints of the medical domain. We emphasize that the AI tools are used solely under human supervision. All AI-generated datasets are rigorously reviewed and validated by the authors for clinical accuracy and ethical compliance. We maintain full responsibility for the final content and the integrity of the published work.

\bibliographystyle{IEEEtran}
\bibliography{mybib}

\end{document}